\documentclass[sigconf]{acmart}

\copyrightyear{2025}
\acmYear{2025}
\setcopyright{rightsretained}
\acmConference[KDD '25]{Proceedings of the 31st ACM SIGKDD Conference on Knowledge Discovery and Data Mining V.2}{August 3--7, 2025}{Toronto, ON, Canada}
\acmBooktitle{Proceedings of the 31st ACM SIGKDD Conference on Knowledge Discovery and Data Mining V.2 (KDD '25), August 3--7, 2025, Toronto, ON, Canada}
\acmDOI{10.1145/3711896.3737185}
\acmISBN{979-8-4007-1454-2/2025/08}

\usepackage{graphicx} 
\usepackage{algorithm}
\usepackage{algpseudocode}
\usepackage{makecell}
\usepackage{xcolor}
\usepackage{stfloats}


\acmConference[KDD '25]{Proceedings of the 31st ACM SIGKDD Conference on Knowledge Discovery and Data Mining V.1}{August 3--7, 2025}{Toronto, ON, Canada}


\title{A Scalable and Efficient Signal Integration System for \\ Job Matching}

\author{Ping Liu}
\affiliation{
 \institution{LinkedIn Corporation}
 \city{Mountain View}
 \state{CA}
 \country{USA}
}
\email{piliu@linkedin.com}

\author{Rajat Arora}
\affiliation{
 \institution{LinkedIn Corporation}
 \city{Mountain View}
 \state{CA}
 \country{USA}
}
\email{rajarora@linkedin.com}

\author{Xiao Shi}
\affiliation{
 \institution{LinkedIn Corporation}
 \city{Mountain View}
 \state{CA}
 \country{USA}
}
\email{xishi@linkedin.com}

\author{Benjamin Hoan Le}
\affiliation{
 \institution{LinkedIn Corporation}
 \city{Mountain View}
 \state{CA}
 \country{USA}
}
\email{ble@linkedin.com}

\author{Qianqi Shen}
\affiliation{
 \institution{LinkedIn Corporation}
 \city{Mountain View}
 \state{CA}
 \country{USA}
}
\email{qishen@linkedin.com}

\author{Jianqiang Shen}
\affiliation{
 \institution{LinkedIn Corporation}
 \city{Mountain View}
 \state{CA}
 \country{USA}
}
\email{jershen@linkedin.com}

\author{Chengming Jiang}
\affiliation{
 \institution{LinkedIn Corporation}
 \city{Mountain View}
 \state{CA}
 \country{USA}
}
\email{cjiang@linkedin.com}

\author{Nikita Zhiltsov}
\affiliation{
 \institution{LinkedIn Corporation}
 \city{Mountain View}
 \state{CA}
 \country{USA}
}
\email{nzhiltsov@linkedin.com}

\author{Priya Bannur}
\affiliation{
 \institution{LinkedIn Corporation}
 \city{Mountain View}
 \state{CA}
 \country{USA}
}
\email{pbannur@linkedin.com}

\author{Yidan Zhu}
\affiliation{
 \institution{LinkedIn Corporation}
 \city{Mountain View}
 \state{CA}
 \country{USA}
}
\email{yidzhu@linkedin.com}

\author{Liming Dong}
\affiliation{
 \institution{LinkedIn Corporation}
 \city{Mountain View}
 \state{CA}
 \country{USA}
}
\email{mdong@linkedin.com}

\author{Haichao Wei}
\affiliation{
 \institution{LinkedIn Corporation}
 \city{Mountain View}
 \state{CA}
 \country{USA}
}
\email{hawei@linkedin.com}

\author{Qi Guo}
\affiliation{
 \institution{LinkedIn Corporation}
 \city{Mountain View}
 \state{CA}
 \country{USA}
}
\email{qguo@linkedin.com}

\author{Luke Simon}
\affiliation{
 \institution{LinkedIn Corporation}
 \city{Mountain View}
 \state{CA}
 \country{USA}
}
\email{lsimon@linkedin.com}

\author{Liangjie Hong}
\affiliation{
 \institution{LinkedIn Corporation}
 \city{Mountain View}
 \state{CA}
 \country{USA}
}
\email{liahong@linkedin.com}

\author{Wenjing Zhang}
\affiliation{
 \institution{LinkedIn Corporation}
 \city{Mountain View}
 \state{CA}
 \country{USA}
}
\email{wzhang@linkedin.com}

\renewcommand{\shortauthors}{Ping Liu et al.}
\renewcommand{\shorttitle}{LinkedIn STAR System}

\begin{document}

\begin{abstract}

LinkedIn, one of the world’s largest platforms for professional networking and job seeking, encounters various modeling challenges in building recommendation systems for its job matching product, including cold-start, filter bubbles, and biases affecting candidate-job matching. To address these, we developed the \textit{STAR} (\textbf{S}ignal Integration for \textbf{T}alent \textbf{a}nd \textbf{R}ecruiters) system,  leveraging the combined strengths of Large Language Models (LLMs) and Graph Neural Networks (GNNs). LLMs excel at understanding textual data, such as member profiles and job postings, while GNNs capture intricate relationships and mitigate cold-start issues through network effects. \textit{STAR} integrates diverse signals by uniting LLM and GNN capabilities with industrial-scale paradigms including adaptive sampling and version management. It provides an end-to-end solution for developing and deploying embeddings in large-scale recommender systems. Our key contributions include a robust methodology for building embeddings in industrial applications, a scalable GNN-LLM integration for high-performing recommendations, and practical insights for real-world model deployment.

\end{abstract}

\maketitle

\section{Introduction}
\label{sec.intro}

LinkedIn is the leading professional social networking platform, primarily focused on recruitment and employment. It offers a space for career networking, professional growth, job recruitment, and more. Users often connect with others who share common experiences, such as education, work history, or similar professional backgrounds. These connections are valuable for advancing careers, discovering new opportunities, and enhancing professional skills. The platform contains rich textual information, including member profiles, job postings, and company pages, which help members explore opportunities and connect. In essence, LinkedIn is designed to cultivate an environment that promotes career development and professional networking, particularly for job seekers.

Traditional recommender systems face challenges such as cold-start, filter bubbles, bias and fairness \cite{liu2021interaction,liu2024does}. In the talent and recruiter marketplace, these challenges evolve with context. For example, the cold-start issue typically refers to users who have not been actively seeking job opportunities in recent periods. Filter bubbles may limit job seekers to specific industries, preventing them from discovering opportunities outside their usual scope. Additionally, candidates recommendations require a higher level of accuracy and relevance, factoring in education, seniority, work experience, and more. Unlike other recommendation systems, job posters often expect a small pool of highly qualified candidates, which makes precise matching critical for filling positions effectively \cite{LinkRetrieval2024}. 

Embeddings have proven to be a key building block for recommender systems, as pre-training on large volume of data can enhance downstream models' resilience to the above mentioned challenges, enable efficient training, and lower real-time computation cost.
Large Language Models (LLMs) have made significant advances in natural language understanding tasks, particularly in understanding long-sequence text such as member profiles and job postings. 
GNNs have proven effective in addressing cold-start problems by allowing common and relational information to flow through connected edges. They are effective at capturing complex relationship, including user preferences, item similarities, and contextual information. Previous studies show that people with shared skills can explore jobs across different industries more efficiently by leveraging GNN embeddings \cite{liu2024linksage}. GNNs have also shown strong results in industry applications, particularly in modeling complex relationships that directly impact business metrics \cite{el2022twhin,borisyuk2024lignn}.

In this paper, we propose the \textit{STAR} (\textbf{S}ignal Integration for \textbf{T}alent \textbf{a}nd \textbf{R}ecruiters) system, which combines the strengths of LLMs and GNNs. We believe LLMs and GNNs work in synergy to provide most crucial signals to any recommendation engine. Where LLMs excel at encoding textual information, GNNs make use of network effects, thereby enhancing the use of semantic information within the application’s context. Without LLMs, graphs would struggle to encode nuanced semantic information whereas LLMs alone would ignore realtime user interactions and feedbacks. Thus, we introduce an end-to-end scalable solution that demonstrates how to effectively train, deploy, and serve LLM embeddings within an industrial recommender system. To enhance the performance of AI models across different business domains, we leverage GNNs to integrate diverse signals -- such as nodes, edges, and input features -- from multiple business products. Through both offline and online A/B tests, \textit{STAR} has shown significant positive business impacts across three distinct models. Our key contributions are as follows:


\begin{itemize}
    \item We presented a comprehensive handbook on training and deploying LLM embeddings with long sequences for industrial applications, which reduced the pain points of categorical features  and provide better improved representation. 

    
    \item We introduced the \textit{STAR} system, which integrates GNNs and LLMs, enabling it to support multiple business models. The proposed approach boosts productivity and reduces operational cost. Our system is not only providing model training techniques, but also the practical solution, such as adaptive sampling and version management.  

    \item Our experiments demonstrated the effectiveness of the proposed framework, showcasing enhanced user experience through significant improvements in offline evaluations and substantial business impact across three different products.
\end{itemize}

The rest of the paper is organized as follows: Section~\ref{sec.related_work} reviews related work, followed by our approach to LLM embeddings in Section~\ref{sec.methodology_llm}. The design of the \textit{STAR} system is detailed in Section~\ref{sec.star}. Section ~\ref{sec.experiment} provides an analysis of offline and online experimental results, and discusses our design choice, as well as the limitations. Finally, we conclude in Section ~\ref{sec.conclusion}.




\section{Related Work}
\label{sec.related_work}

\textbf{Graph Neural Network (GNN)}. GNN~\cite{zhou2020graph} is designed to process and model complex data structures that can be represented as graphs~\citep{liu2023graphprompt,yang2024graphpro,yu2024multigprompt}. A lot of research work focuses on how to handle and aggregate the dynamic neighbors information on the modeling side ~\cite{hamilton2017inductive, kipf2016semi, kipf2016semi, velivckovic2018graph}. Recently, GNN has seen increasing adoption across various industries due to its power to handle complex relational data. Twitter~\cite{el2022twhin} utilizes knowledge-graph embeddings for entities within Twitter's network to help ad ranking, follower recommendation, search ranking and other scenarios. Pinterest~\cite{ying2018graph} leverages a random walk strategy to structure convolution graphs and a progressive strategy to handle harder examples to improve model robustness and convergence.~\citet{damianou2024towards} introduces a Heterogeneous GNN that captures multi-hop relationships between different types of recommendable items.

\textbf{LLM for Recommendation}. Integrating LLMs into recommender systems has become a key research focus, with studies divided into two approaches based on LLM application methods. The first approach uses LLM-based embeddings as features within recommender models, enriching user and item representations with semantic information extracted from unstructured data sources such as reviews, user-generated content, and item descriptions~\citep{wu2024unify}. For instance,~\citet{zhao2021embedding} constructed an embedding-based recommender system for job to candidate matching,~\citet{shin2021one4all} developed ShopperBERT that learns user embeddings for E-commerce recommendation, and~\citet{hou2022towards} proposed a universal sequence representation learning approach to learn transferable representations of associated description text of items. 
The second approach focuses on developing the next generation of recommender systems, where LLMs themselves function as the core engine. ~\citet{hou2024large} demonstrated that existing LLMs can be viewed as zero-shot rankers, which can sort the relevance of movies based on user historical interactions and movie descriptions, ~\citet{bao2023tallrec} built TALLRec, a large recommendation language model based on LLaMA-7B~\citep{touvron2023llama} by tuning LLMs with recommendation data,~\citet{zhu2024collaborative} proposed CLLM4Rec, the first generative recommender system that tightly couples the ID paradigm and the LLM paradigm.

\textbf{Combining LLMs and GNNs}. There is some early research on the combination of LLMs and GNNs~\citep{pham2023deep}, driven by the complementary strengths of these models. LLMs excel in understanding complex user-item interactions expressed in natural language, while GNNs are particularly effective at capturing structural and relational information. Research can be broadly categorized into two approaches. 
The first approach uses LLMs to create textual embeddings that are then aggregated by GNNs. For example, GraphSage \cite{hamilton2017inductive} encodes node features independently before aggregating them with GNNs to produce final node representations -- a method widely adopted by later works~\citep{zhutransformer2021,hu2020gpt,li2021adsgnn,wang2023learning}.
The second approach involves co-training LLMs and GNNs together. For example, \citet{yang2021graphformers} introduced GraphFormers, which interleave GNN components with transformer layers, enabling text encoding and graph aggregation to work together. Similarly, \citet{xie2023graph} proposed the graph-aware language model pre-training to integrate LLMs with GNNs, enhancing downstream applications when fine-tuned. \citet{choudhary2024interpretable} introduced the Plug and Play Graph Language Model to combine LLM and GNN predictions using a Gradient-Boosted Tree for improved query-product matching.

While our work is similar to previous approaches that train LLM embeddings and GNNs, we face distinct challenges and scale requirements for our products. In the literature discussed above, the LLM encoder (or embedding model) typically employs a BERT-like architecture~\citep{devlin2019bert}, which can handle only limited sequence lengths (usually no more than 512 tokens). Job postings and member profiles, however, are substantially longer, averaging between 2000 and 3000 tokens. To address this, we selected Mistral-7B-Instruct~\citep{jiang2023mistral7b, wang2023improving}, a model with 7 billion parameters -- significantly more than BERT’s 110 million~\citep{devlin2019bert}. Co-training such large models would typically require thousands of GPUs, making it prohibitively expensive at our scale. We present an end-to-end solution for developing and deploying embeddings at scale, requiring only a handful of GPUs for training. While we use the job matching product to illustrate the problem and applications, the solution is general enough for any modern relevance system. The shared insights will be valuable to other ML practitioners in the industry.

\section{Large Language Model Embedding}
\label{sec.methodology_llm}

While using LLMs for text embedding is not a new concept, it is often limited to processing specific chunks of the input. Many models heavily rely on categorical features, which is usually inferred from a maintained taxonomy. For example, a user's current job title is first linked to the most appropriate Title ID in the taxonomy database, which is then used as input for the ranking model. This approach has two main drawbacks: (1) ID-based features often rely on Named Entity Recognition models, which can be complex to train, introduce operational overhead, and are prone to errors, and (2) taxonomies can be incomplete, difficult to update, and assume a one-size-fits-all approach, missing important nuances. 
Table~\ref{tab.case_study} shows examples of these challenges in job matching. With the advent of RoPE positional embeddings \cite{su2024roformer} and enhanced algorithms for handling longer context lengths, we propose using the full raw text with minimal modifications as input to generate embeddings directly from an LLM.

\begin{table*}[tb]
{
    \caption{Challenges in categorical feature processing}
    \begin{tabular}{l|c}
    \hline
        Category &  Example Keywords \& Their Issue \\ \hline 
        \midrule
        Seniority & \textit{Managing Director} -- Is it a Director level or VP level? \\ \hline
        Industry & \textit{Bloomberg} -- Financial Service, Media, Technology, or Software Development? \\ \hline
        Taxonomy & \textit{GenAI} -- What is the corresponding skill ID in taxonomy? \\ \hline
        Title & \textit{LLM Tutor} -- No perfect match in an outdated taxonomy  \\  
    \hline
    \end{tabular}
    \label{tab.case_study}
}    
\end{table*}

While zero-shot performance generally improves with model size \cite{DBLP:journals/corr/abs-2001-08361}, modern recommendation systems often serve billions of users and index hundreds of millions of active items, making large models impractical for deployment at scale. Training such models can be prohibitively expensive due to high infrastructure cost and extended training time. Moreover, fine-tuned smaller models have demonstrated superior performance compared to vanilla flagship models \cite{bucher2024finetunedsmallllmsstill}.
Motivated by these insights, we fine-tuned the E5-Mistral-7b-instruct model \cite{jiang2023mistral7b, wang2023improving}, as detailed in the following subsections. We selected Mistral-7b-instruct for two main reasons: (1) Mistral-7b is a robust multilingual model capable of encoding long sequences, pretrained on a vast corpus; and (2) it was fine-tuned for embedding use cases using synthetic data, enabling strong performance with relatively low effort.
While we focus on Mistral-7b, our approach is highly adaptable and can be applied to both smaller and larger models, depending on hardware budget and latency constraints.

\subsection{Modeling Objectives}
We follow a bi-encoder \cite{thakur2021augmented} approach where the job posting forms the first tower, dubbed  ``document'' and the member profile and resume together form the other tower, dubbed ``requests''. We chose a bi-encoder over a cross-encoder \cite{reimers2019sentence} due to (1) higher throughput of bi-encoder, as the attention mechanism has a complexity of $O(n^2)$ with respect to sequence length $n$; and (2) easier deployment, as discussed later in Section~\ref{sec.model_serving}.

The model is fine-tuned for binary classification, where a label of $1$ indicates that a member applied to a job, and $0$ otherwise. Job applications generally reflect a stronger alignment between the job and the member: users typically consider multiple factors before applying, such as qualifications, experience fit and career progression, whereas a simple click may indicate only initial interest. To enhance performance, we augment binary cross-entropy (BCE) loss with triplet loss \cite{chechik2010large, alpaydin2020introduction} after extensive experiments, selecting the semi-hard \cite{DBLP:journals/corr/SchroffKP15} in-batch sample as the negative. Using semi-hard negatives, rather than the hardest negatives or the entire batch, proved advantageous as it reduces false negatives. For each document sample, we mine the semi-hard negative sample from both components of the request tower, resulting in the optimization objective below:

\begin{equation}
L = L_{bce} + \lambda L_{c}
\end{equation}
where binary cross entropy (BCE) term is defined as
\begin{equation}
    L_{bce} = - \left[ y \log(y_{\text{pred}}) + (1 - y) \log(1 - y_{\text{pred}}) \right]
\end{equation}
and contrastive term is defined as

\begin{equation}
    L_{c} = -\sum_{i \in \mathcal{B}, \, y_i = 1} \log \left( \frac{\exp(\text{sim}(z_d, z_r) / \tau)}{\exp(\text{sim}(z_d, z_r) / \tau) + \exp(\text{sim}(z_d, z_r^-) / \tau)} \right)
\end{equation}
In the equations above \( y \) and \( y_{pred} \) are actual and predicted label values, \(z_d \) and \(z_r \) are document and request embeddings respectively and  \(z_r^- \) is semi-hard request embedding and  $\tau$ is the temperature parameter and $\text{sim}$ is the similarity function which is dot product for our experiments.  \(z_r^- \) is chosen such that 

\begin{equation}
z_r^- = \arg\max_{z_{r'} \in \mathcal{B} \setminus \{z_r\}} \left\{ \text{sim}(z_{r'}, z_d) \, \big| \, \text{sim}(z_{r'}, z_d) < \text{sim}(z_r, z_d) \right\}
\end{equation}

It’s important to note that the same model is used to embed all jobs, member profiles, and resumes, significantly reducing both serving and training cost. The inputs are distinguished using an input prefix-prompt specifying the nature of input. The final output embedding is average of <CLS> token across all layers followed by  $L_2$ normalization. For training the model we calculate pairwise hadamard product between inputs to document and request towers followed by linear layers to predict \( y_{pred} \).

\subsection{Model Training}
We adopted LoRA \cite{hu2021lora} with a rank of 8 to fine-tune models, which supports faster convergence and allows for larger batch sizes. We further optimize training by using mixed precision with bfloat16. Using tower specific LoRA adapters didn't lead to any major improvement as compared to using same LoRA adapters for all the towers, hence we use one set of adapters. We use a maximum context length of 1800 tokens for job descriptions and member resumes, and 1024 tokens for member profiles. This context size is memory-intensive, and despite using PEFT \cite{peft} with LoRA and bfloat16, we were limited to a batch size of 16 on a single A100 GPU. Such small batch size can cause inefficient training due to inconsistent gradients. To increase the per-device batch size, we enable gradient checkpointing to free up memory by discarding layer activations during the forward pass and recalculating them as needed during the backward pass. Using gradient checkpointing (hence a higher batch size) consistently yields better results compared to lowering the batch size and training for an equivalent duration.

Even with gradient checkpointing, the per-device batch size may still be insufficient to produce consistent gradients, making gradient accumulation necessary to reach an adequate effective batch size. We define effective batch size as follows:

\begin{equation}
    num\_gpu  \times per\_device\_batch\_size \times grad\_accumulation\_steps
\end{equation}

We observe consistent improvements by increasing the batch size to 3172. Further tuning, however, does not yield better offline metrics. The model is trained using the AdamW optimizer \cite{loshchilov2017decoupled}. To enhance performance, the number of warm-up steps is adjusted inversely with the effective batch size. Specifically, for a batch size of 3172, the warm-up steps are set to 30.

\subsection{Training Data}
\label{sec.training_data}
We use member activities as the source of our training data, following the industry standard. 
The data we use is managed in accordance with relevant member settings to ensure full respect for their privacy choices.
Although job applications are  16x rarer than skips or clicks, we maintain a 1:1 ratio between positive and negative samples in our training dataset. This adjusted ratio differs from the real-world distribution but allows us to complete training within a 2-day window, due to resource constraints and the need for faster iterations. With this timeframe, we are limited to 4.5 million samples. In this setup, we empirically found that having a higher proportion of positive samples contributes more effectively to model performance than increasing the number of negative samples, hence the decision to keep a balanced dataset. Reweighing the negative samples in loss function to enforce the real world class ratio didn't lead to improvement. 
We solve time-travel issues by using the snapshot values of texts at the moment of the job action.

\subsection{Model Serving}
\label{sec.model_serving}
Text inputs are processed with Apache Beam over Kafka streams. A Kafka event is triggered whenever a job post or member profile is updated. We store an MD5 hash of text records and regenerate embeddings only if the current hash differs from the previous one, to save GPU computation.  
We utilize a cluster of multi-instance A100 GPU \footnote{https://www.nvidia.com/en-us/technologies/multi-instance-gpu/} for LLM inference. The generated embeddings are stored in a key-value store for online usages and sent to a downstream Kafka flow for offline training purposes. Our system currently supports over 200 peak QPS with an end-to-end latency of $\sim$270 ms.

\begin{figure*}[!ht]
    \centering
    \includegraphics[width=0.96\textwidth]{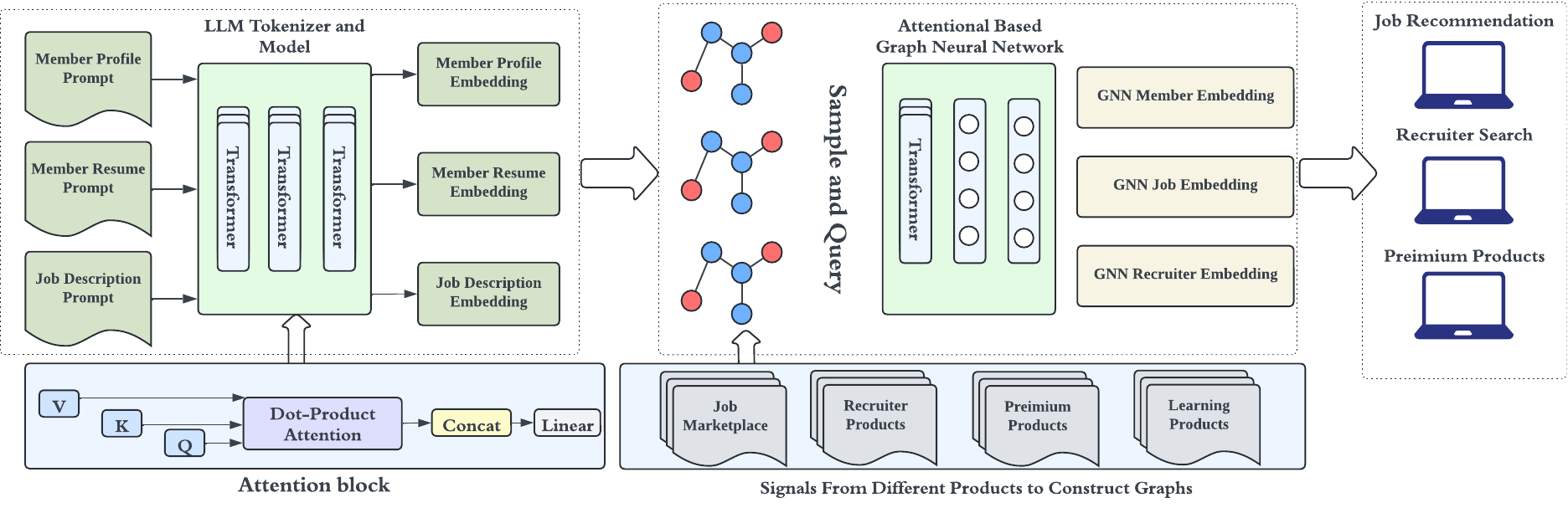}
    \vspace{-6pt}
    \caption{The illustration of the \textit{STAR} workflow. The figure's left side depicts the LLM workflow, which generates embeddings for member profiles, resumes, and job descriptions. The GNN then processes these embeddings during the signal integration stage to support LinkedIn's downstream products.}
    \label{fig.star_pipeline}
    \Description[The pipeline of star.]{The pipeline of star.}
\end{figure*}

\section{Signal Integration for Talent and Recruiter}
\label{sec.star}


LinkedIn maintains a variety of models supporting different business areas, including Premium memberships, Jobs, Recruiter, and Learning. Our vision is to unify these efforts under a more cohesive modeling stack for job matching. Currently, we manage multiple machine learning models that connect LinkedIn members with opportunities such as jobs, recruiters, and courses. Many of the same signals are valuable across all these models, especially those that represent a member's expertise and experience. However, new signals are often integrated piecemeal, starting in one application and gradually spreading to others.

To boost engineering efficiency, \textbf{S}ignal Integration Platform for \textbf{T}alent \textbf{a}nd \textbf{R}ecruiter (\textit{STAR}) was developed to support 
``horizontal-first'', rapid, cross-functional model development across business domains.
Using our GNN infrastructure, \textit{STAR} marked a shift in our feature engineering process, where engineers from multiple teams added relevant signals to the same graph model from the start. Through a branch-based development model, each engineer could independently test and measure the impact of their signal contributions before unifying these branches into a consolidated graph version for release. Once the new graph version was deployed, downstream application models could be retrained to capture full impact. \textit{STAR} has proven effective in enhancing engineering productivity, reducing operational cost, and achieving business gains by leveraging GNNs and LLMs for feature engineering.


\subsection{Graph Design}

\begin{figure}[tb]
    \centering
    \includegraphics[width=0.47\textwidth]{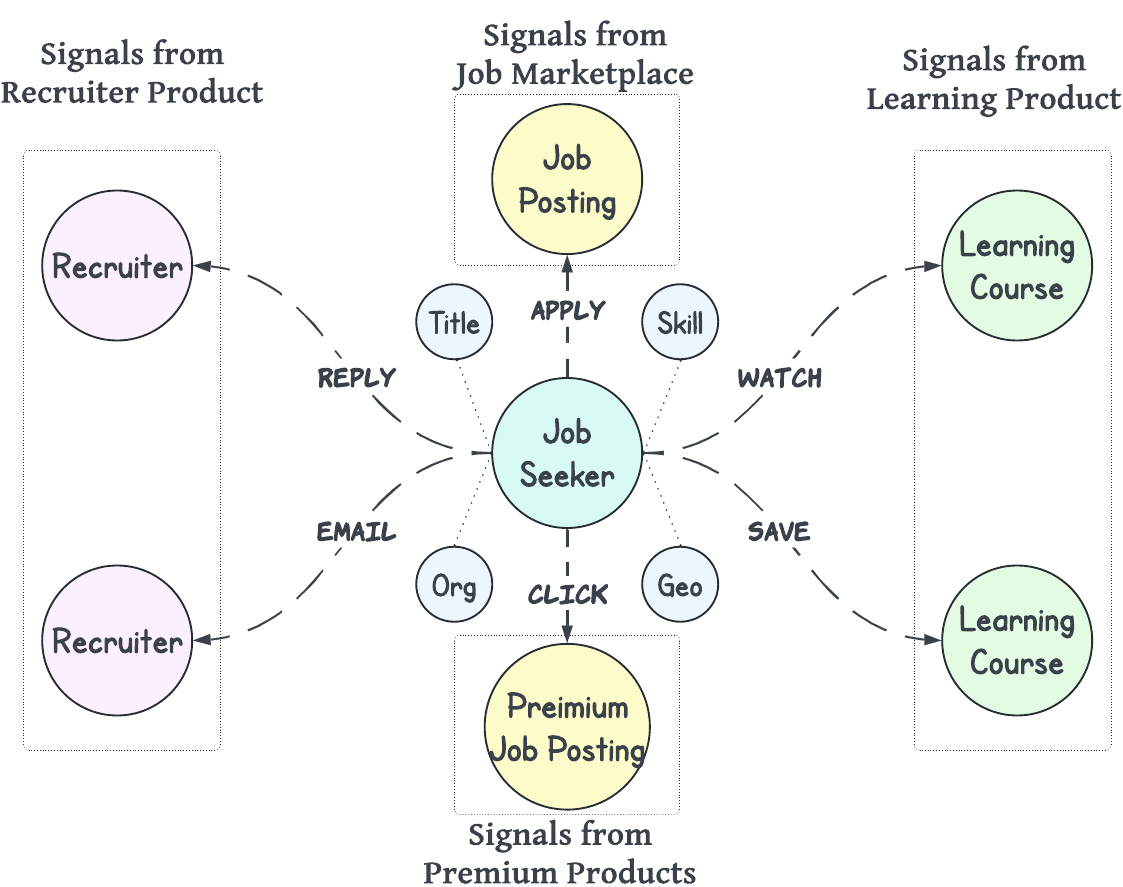}
    \caption{An example of LinkedIn's heterogeneous graph.}
    \label{fig.star_graph}
    \Description[An example of LinkedIn's heterogeneous graph.]{An example of LinkedIn's heterogeneous graph.}
\end{figure}

Given LinkedIn's role as a professional social media platform, constructing a graph with various entities on it is straightforward. This graph consists of various node entities, including members, jobs, recruiters, and learning courses. It also incorporates attribute nodes related to recruiters and jobs, such as geographic location, title, company, seniority, skills, and others.

LinkedIn's graph encompasses two categories of edge types: interaction edges and attribute edges. \textbf{Interaction edges} represent historical activities and interactions between pairs of nodes. Examples include a member \textit{A} applying to a job \textit{B}, a member completing a course on LinkedIn Learning, or a recruiter receiving and accepting an InMail from a job candidate. These edges effectively capture node activities and facilitate message passing across the graph. \textbf{Attribute edges} are the edges connecting entity nodes with attributes nodes. The attributes are selected based on two criteria: (a) they should serve as key factors influencing interactions, and (b) they should be common across different node types. For instance, \textit{skills} are a shared and critical attribute for members, jobs, recruiters, and learning courses. This commonality allows skill-based connections to unify members, jobs, and recruiters in the graph's construction.

There are three types of features that input to each node: (a) LLM embeddings; (b) ID embeddings; (c) categorical features. The details of LLM embedding is discussed in Section~\ref{sec.methodology_llm}, which covers both member profile embedding, and job description embedding. ID embeddings are used for attribute nodes, such as title, company, skill to featurize the nodes. Categorical features are used otherwise, which covers the semantic representation in different aspects. All training data was used in a privacy-preserving manner, and no personally identifiable information (PPI) was included.

\subsection{GNN Training Framework}

We choose to use the open-source DeepGNN \footnote{https://github.com/microsoft/DeepGNN} library as our graph engine for the GNN training framework. Given the size of graph is around 6 Terabytes, to launch and load the graph metadata, we deploy 30 machines where each one has 230 Gigabytes memory. For training, we use 36 v100 Nvidia GPUs to start the training jobs. For inference, using 150 AMD Ryzen7 CPU nodes is able to achieve equivalent performance with 36 v100 GPUs. 

The major bottleneck of training and inference is I/O and neighbor sampling. It shows ~70\% of training time is due to the sampling and I/O operations from graph engine. To speed up the training, we adopt and simplify the adaptive sampling proposed by \citet{borisyuk2024lignn} during the training, which saves one third of the training time and decreases the training job from 48 hours to 36 hours. 

\begin{algorithm}
\small
\caption{Adaptive Sampling} \label{algo:adaptive_neghbor_sampling}
\begin{algorithmic}
    \Require: graph $G(V, E)$; \# maximum sampled neighbors: $\alpha$; \# initial sampled neighbors: $\sigma$; incremental stride: $\delta$.
    \Ensure: A trained model: $M$
    \Function{Adaptive\_Sampling}{}
        \State sample\_count $\gets$ $\sigma$     
        \For{epoch \textbf{in} epochs}
           \For{step \textbf{in} steps}
                \State train\_data $\gets$ query ($G$, sample\_count)
                \State gradient($M$)
            \EndFor 
            \If{evaluate ($M$, val\_data) not improving}
                \State sample\_count $\gets$ min(sample\_count + $\delta$, $\alpha$) 
            \EndIf
        \EndFor
        \State \textbf{return} $M$
    \EndFunction
\end{algorithmic}
\end{algorithm}

\subsection{Model Architecture}

We adopt an encoder-decoder architecture in GNN modeling. The encoder is designed to capture and learn the representation of the respective node entities, while the decoder performs optimization tasks that are explicitly defined and aligned with the requirements of the downstream applications. Our framework supports GraphSAGE \cite{kipf2016semi}, temporal sampling \cite{borisyuk2024lignn}, and personalized PageRank (PPR) sampling \cite{zeng2021decoupling}. These sampling methods can be coupled with Adaptive sampling in Algorithm~\ref{algo:adaptive_neghbor_sampling} to speed up training. 

The encoder is composed of multiple multi-head attention layers, with an option to include the target node in the attention mechanism. The activation function used is LeakyReLU, and both L2 regularization and batch normalization are applied at the end of the encoder layers. The embedding size (which also represents the output dimension of the encoder) is set to 200. Our experiment shows a positive correlation between embedding size and model performance when tested within the range of 50 to 200.

\subsection{GNN with LLM Embedding} 

We explore two training strategies for integrating GNN and LLM models. In the first approach, we combine the LLM encoder with the GNN encoder and either co-train both models or freeze the LLM encoder. We refer to this method as GNN + LLM Encoder. In the second approach, we fine-tune the LLM separately and use its embeddings as node features within the GNN, which we call GNN + LLM Embedding. After comparing performance, computational cost, and training speed, we chose the second approach for our product for several key reasons. First, using pre-computed embeddings allows us to decouple the training of the LLM and GNN, significantly reducing computational overhead. This method also requires less memory and hardwares, as the LLM model is not involved in the backpropagation process during GNN training. Additionally, it offers greater flexibility in handling larger batch sizes and neighbor sampling, enhancing overall training speed and allowing for more adaptable hyper-parameter tuning. As a conlcusion, GNN + LLM Embedding is a more practical and scalable solution for our product given its better performance at a lower cost. Details of the experiments conducted to compare these methods are provided in Section~\ref{sec:GNN-LLM-offline} and Appendix~\ref{sec:gnn+llm-encoder}.

\subsection{Optimization Objectives}

The objectives of GNN training can be typically modeled as either link predictions or node classification problems. It helps in discovering hidden or missing links (edges) between nodes, which is widely used in social networks, recommendation systems, knowledge graphs and transferring learning. In \textit{STAR}, each downstream task is formalized as link prediction, the training objective is to minimize the cross-entropy loss:

\begin{equation}
\begin{split}
 \textit{Loss} = -\sum_{i\in S_s}\sum_{j\in S_d} ( y_{ij} \log(\sigma(\text{score}(i, j))) \\
 + (1 - y_{ij}) \log(1 - \sigma(\text{score}(i, j)))),
\end{split}
\end{equation}

where $S_s$ and $S_d$ are the source nodes and destination nodes sets. To expedite productivity and engineering cost, we set up multi-task learning to handle embeddings for different entities. The tasks are detailed in Section \ref{sec.experiment}. The multi-objective problem is formulated as a multi-loss function:

\begin{equation}
 \textit{Loss} = \sum \lambda_{i} \text{$L_{i}$},
\end{equation}

where $\lambda_i$ is a hyper-parameter that can be tuned in practice, $L_i$ represents the loss of $task_i$ in multi-task learning. 

\subsection{Life Cycle Management of Embeddings}

Embeddings are continuously updated as models evolve, and each new version may be incompatible with previous ones. Due to limited infrastructure capacity, older embedding versions typically need to be deprecated when deploying new ones. To effectively utilize the latest embeddings, downstream models must be retrained, which creates challenges for both producers and consumers. This dependency often extends the migration and deprecation process to months or even quarters.

\textit{STAR} not only accelerates feature engineering productivity but also addresses the challenges of version management. Building on prior work on embedding management \cite{hu2022learning}, we propose a new framework for managing the embedding version lifecycle with backward compatibility —- enabling embeddings that are backward compatible so retraining is not (always) required. 
The objective of backward compatibility is to learn a linear transformation function, as defined below.
\begin{equation}
    B_k^{k-1} = \textit{transform} (E_k) \approx E_{k-1},
\label{eq.backward_1}
\end{equation}

where $B_k^{k-1}$ is backward compatible embedding trained with new version $k$, and backward compatible with version $k-1$; $E_k$ is latest version of embedding with dimension $m$; $E_{k-1}$ is the embedding of the previous version with dimension $n$; \textit{transform} represents a linear transformation function, which could be a simply weight matrix with $n \times m$ dimensions. The loss function can be defined as $L2$-norm loss between $B_k^{k-1}$ and $E_{k-1}$: 

\begin{equation}
    \textit{Loss} = L2_{norm}(B_k^{k-1}, E_{k-1})
\label{eq.backward_loss}
\end{equation}

Both offline and online A/B tests demonstrate that the backward-compatible embedding achieves parity results without any degradation in metrics. Additionally, it only requires maintaining a weight matrix, significantly reducing storage cost, computational hardware requirements, and engineering maintenance effort.

\section{Experimental Results}
\label{sec.experiment}

In this section, we will discuss the details of training task of GNN and LLM. Then we will share our offline evaluation and hyper-parameter selections, as well as the analysis of online A/B tests. 



\paragraph{Job Recommendation \& Search}
Job recommendations aim to promote relevant jobs to seekers based on their profiles, preferences, and past activities. In search, the system interprets the intent of a query and personalizes results further based on member profiles. A positive label indicates that a member has applied to a job, while negatives  include actions ``skip'' or ``dismiss'' on a displayed job. 

\paragraph{Recruiter InMail Messages} Talent Recommendation is a LinkedIn product designed to suggest potential candidates who match the requirements of a specific job. 
We model not only the match between recruiters and candidates but also the likelihood that a candidate will respond to an InMail message. Negative samples are drawn from recruiter-candidate pairs where no reply action occurred. 

\paragraph{Top Applicant Jobs} Top Applicant Jobs is a product for Premium users that aims to enhance engagement between recruiters and Premium job seekers. Positive interactions, such as emails, messages, or phone calls initiated after a job seeker applies, are modeled as positive examples in this task to support business growth.

\textit{STAR} ensures that training data is managed in accordance with relevant member settings, fully respecting their privacy choices. Additionally, de-identification is applied during the training phase to safeguard member confidentiality and privacy.

\subsection{Offline Evaluation}
\label{sec:offline-eval}
\subsubsection {LLM Offline Evaluation}

The baseline is to use a frozen LLM model and trainable linear layers on top to evaluate out-of-the-box performance. As shown in Table~\ref{tab_llm_offline}, frozen E5-Mistral model ~\citep{jiang2023mistral7b, wang2023improving} achieves respectable performance with AUC of $0.6886$. However this is below the threshold for acceptable performance. We further fine-tune it  with binary cross-entropy (BCE) loss \cite{alpaydin2020introduction} to improve performance by $5\%$ to $0.7331$. Further adding contrastive loss and gradient checkpointing provides marginal bump of $0.3\%$ each.


\begin{table}[tb]
\small{
\caption{Offline AUC of LLM training algorithms}
\vspace{-6pt}
\begin{tabular}{l|l}
\hline
Experiment & AUC \\
\hline
\midrule
Vanilla Mistral & 0.6886 \\
\hline
+ fine-tune with BCE & +4.5\% (0.7331) \\
\hline
+ Contrastive & +0.3\% (0.7363)\\
\hline
+ Gradient Checkpointing & +0.3\% (0.7385) \\
\hline
\end{tabular}
\label{tab_llm_offline}
}
\end{table}

\subsubsection{GNN + LLM Offline Evaluation}
\label{sec:GNN-LLM-offline}

\begin{table*}[tb]
    \centering
    \small{
    \caption{GNN model performance on validation set with different hyper-parameter configurations}
    \label{tab:hyperparameters}
    \vspace{-6pt}
    \scalebox{0.85}{
    \begin{tabular}{l|l|l|l|l|l|l|l|l}
    \toprule
        \makecell[l]{Encoder \\ Dimension} & \makecell[l]{Encoder Node \\ Units Multiplier} & Sampling & Pooling & \makecell[l]{Dual \\ Encoder} & \makecell[l]{Encoder \\ Activation} & \# 
        \makecell[l]{\# Training \\ Instances} & \makecell[l]{\# Validation \\ Instances}  & Val AUC \\ \midrule \midrule
        100 & 1 & Random & mean & No & Linear & 300M & 30M & 0.8215 \\  \hline
        100 & 1 & Random & mean & No & Linear & 500M & 30M & 0.8258 \\  \hline
        200 & 1 & Random & mean & No & Linear & 500M & 30M & 0.8341 \\  \hline
        200 & 1 & PPR 50 5 & mean & No & Linear & 500M & 30M & 0.8338 \\ \hline
        200 & 1 & PPR 200 5 & mean & No & Linear & 500M & 30M & 0.8336 \\  \hline
        200 & 1 & Random & mean & Yes & Linear & 500M & 30M & 0.8347 \\ \hline
        200 & 2 & Random & mean & Yes & Linear & 500M & 30M & 0.8363 \\ \hline
        200 & 4 & Random & self-attention & Yes & LeakyRelu & 500M & 30M & 0.8378 \\  \hline
        200 & 4 & Random & self-attention & Yes & LeakyRelu & 800M & 30M & 0.8424 \\ \hline
        200 & 4 & Random & mean & Yes & LeakyRelu & 1B & 30M & 0.8450 \\ \hline
        \textbf{200} & \textbf{4} & \textbf{Random} & \textbf{self-attention} & \textbf{Yes} & \textbf{Leaky Relu} & \textbf{1B} & \textbf{30M} & \textbf{0.8470} \\ 
        \bottomrule
    \end{tabular}}}
\end{table*}

\begin{table*}[tb]
    \small{
    \caption{Performance and resource utilization of GNN models with and without LLM embeddings during offline training}
    \label{tab:gnn-llm-embedding-training}
    \vspace{-6pt}
    \scalebox{0.9}{
    \begin{tabular}{l|l|l|l|l|l|l|l}
    \toprule
        \makecell[l]{Model} & Train Loss & Val Loss & Val AUC & \makecell{\# Trainable \\ Parameters} & \makecell{Hours/ \\ 0.1 epoch} & CPU Memory & GPU Memory \\ \hline \midrule
        \makecell[l]{ ~ Baseline ~} & 0.4048 & 0.4008 & 0.8447 & 12,176,650 & 3.5 & 30GB & 8GB \\ \hline
        \makecell[l]{~ + LLM embedding  ~ } & 0.4019 & 0.3964 & 0.8489 & 12,995,850 & 5 & 60GB & 16.7GB \\ \hline
        \makecell[l]{+ LLM embedding \\ - categorical features  ~} & 0.4092 & 0.4030 & 0.8413 & 6,181,630 & 4 & 50GB & 9GB \\ \hline
        \makecell[l]{+ LLM embedding \\ - legacy embedding  ~} & 0.4010 & 0.3959 & 0.8481 & 12,955,850 & 5 & 60GB & 16.7GB \\ \hline
        \makecell[l]{+ LLM embedding \\ - categorical features \\ - legacy embedding} & 0.4095 & 0.4051 & 0.8398 & 6,141,630 & 4 & 50GB & 8GB \\ 
        \bottomrule
    \end{tabular}}
    \captionsetup{width=.75\textwidth}
}   
\end{table*}

\paragraph{Graph and Dataset}

We construct our graph using six months of LinkedIn Talent Marketplace activity data, resulting in a training graph with 763 million nodes and 12.3 billion edges. The distributions of node and edge types are provided in Table~\ref{tab:nodes-gnn-llm} and Table~\ref{tab:edges-gnn-llm} in the Appendix. For model training, we use 1.5 months of data, while the testing phase is conducted on a 15 days dataset. The training and testing dataset sizes for the three relevance tasks -- \textit{Job Recommendation \& Search}, \textit{Recruiter InMail Messages}, and \textit{Top Applicant Jobs} -- are (1B, 30M), (84M, 12.5M), and (114M, 25M), respectively.


\paragraph{GNN + LLM Embedding} 

We integrated LLM embeddings (e.g., job embeddings) from Section~\ref{sec.methodology_llm} as node features by concatenating them with other relevant features for each corresponding node. To assess the impact of these embeddings, we created several model variants to evaluate the performance of GNN models with LLM embeddings. Our baseline model uses a GNN that incorporates categorical features and legacy embeddings —- 200-dimensional dense embeddings pre-trained using LinkedIn’s two-tower model, which predicts click-through rates using categorical features,  a BERT ~\citep{devlin2019bert} and a T5-small~\citep{chung2024scaling} text encoder. We then explored model variants that added LLM embeddings, removed 22 categorical features, or removed the legacy embeddings.

We used the full training and testing datasets described above for training and validation in each epoch, with validation checks every 0.1 epoch. We used a learning rate of 1e-4 and sampled up to 100 neighbors. Table~\ref{tab:gnn-llm-embedding-training} presents the performance of these models on the validation set, along with resource utilization across different model variants, with all hyper-parameters held constant.

It is shown that incorporating LLM embeddings into GNN model improves performance. For example, the validation AUC increased from 0.8447 (baseline) to 0.8489 when LLM embeddings were included alongside both standardization and legacy embeddings. Even when categorical features or legacy embeddings were removed, models with LLM embeddings maintained competitive performance, with AUCs of 0.8413 and 0.8481, respectively. This demonstrates that LLM embeddings capture critical information for the task. Although incorporating LLM embeddings results in higher resource demands including increased memory usage 
and longer training times, when replacing 22 categorical features, GPU cost remains nearly identical to the baseline, with CPU cost impact being negligible. LLM embeddings offer practical advantages, such as reducing the need for maintenance and updates compared to hand-crafted categorical features. 

Table~\ref{tab:hyperparameters} summarizes the impact of various hyper-parameters on model performance. In our experiments, the learning rate is fixed at 1e-4, the batch size at 4096, epoch at 1, and both batch normalization and L2 regularization were applied. We observe that increasing the encoder dimension, encoder node multiplier, or the size of the training set results in notable performance gains. The best performance is achieved with 4x encoder node multiplier, self-attention pooling, a dual encoder architecture, and LeakyReLU activation, with 1 billion training instances yielding a validation AUC of 0.8470. Incorporating PPR sampling \cite{zeng2021decoupling} does not lead to improvement, and the AUC even decreases slightly to approximately 0.8338 compared to 0.8341 under the same conditions without PPR sampling.

The models were rigorously evaluated for bias and fairness before being released to the public. The final model, incorporating these features, was evaluated across gender, job-seeking urgency, seniority, and location dimensions. All cohorts were within an acceptable range of baseline performance as measured by the AUC of pApply. For example, in the gender dimension, we see a similar AUC between the FEMALE\_Urgent group and the MALE\_Urgent group.

\subsection{Online A/B Tests}
\label{sec:online-eval}
As outlined in the previous section, we have three distinct tasks aligned with the business goals of three different products. We conducted online A/B tests for each task, using a 50\% vs. 50\% traffic split among members. Data was collected over a period ranging from two to twelve weeks and thoroughly reviewed by experts. All reported metrics underwent a two-tailed t-test, with $p$-values less than 0.05, indicating statistical significance. 

\begin{table}[tb]
\small{
\caption{Online metrics in job recommendation \& search. }
\label{tab.job.online}
\vspace{-6pt}
\begin{tabular}{@{\hskip 20pt}l@{\hskip 8pt}|l}
\toprule
Metric Name &  Relative Ratio \\
\midrule \midrule
Dismiss To Apply Ratio         & $-3.2\%$  \\ \hline
Total Applies                  & $+1.5\%$  \\ \hline
Successful Job Search Sessions & $+1.0\%$  \\       
\bottomrule
\end{tabular}
}
\end{table}

\paragraph{\textbf{Job Recommendation \& Search}}

We conducted A/B tests on job recommendation \& job search simultaneously, using the same treatment and control groups. Shown in Table \ref{tab.job.online}, we observed $+1.5\%$ in the number of job applications site-wide. 
We also saw $-3.2\%$ in the dismiss-to-apply ratio, indicating that the GNN variant not only facilitated more applications but also reduced dismissals from users. For the job search product, one of the key metrics is a successful job search session, defined by whether a user applies, saves, or creates a job alert during the session. We observed a 1\% site-wide increase in this metric, reflecting improved search outcomes.

\paragraph{\textbf{Recruiter InMail Messages}} 

In the recruiter product, the key business metrics include InMail acceptance, InMail reply, and overall InMail acceptance rate. Data collected over an eight-week period, summarized in Table \ref{tab.recruiter}, reflects metrics averaged across 7-day intervals. We observed a notable increase in interactions between candidates and recruiters, with most candidates responding within the first three days. Additionally, with a 3\% increase in InMail acceptance, over 85\% of those who accepted went on to reply to recruiters, demonstrating even stronger engagement signals.

\begin{table}[tb]
\small{
\caption{Online metrics in Recruiter InMail Messages.}
\label{tab.recruiter}
\vspace{-6pt}
\begin{tabular}{@{\hskip 20pt}l@{\hskip 8pt}|l}
\toprule
Metric Name &  Relative Ratio \\
\midrule \midrule
Sender InMails Accepted 7d & +3.0\% \\ \hline
Sender InMails Replied 7d & +2.7\% \\ \hline
Sender InMails Accepted 7d Rate & +1.9\% \\
\bottomrule
\end{tabular}
}
\end{table}

\paragraph{\textbf{Top Applicant Jobs}}

Top Applicant Jobs is designed to optimize interactions between recruiters and Premium users after an application is submitted, with the ``hearing back'' rate -- indicating how often applicants receive responses from recruiters -- serving as the primary business metric. Table \ref{tab.taj} presents A/B test results collected over a 12-week period. The online business metrics closely align with the objectives of GNN training, showing an increase of $+2.4\%$ in positive ratings and a decrease of $-5.5\%$ in rejection rates. Additionally, we observed an increase in job impressions from the top funnel and $+0.29\%$ in the number of sessions.

\begin{table}[tb]
\small{
\caption{Online metrics in Top Application Jobs. }
\vspace{-6pt}
\begin{tabular}{@{\hskip 20pt}l@{\hskip 8pt}|l}
\toprule
Metric Name &  Relative Ratio \\
\midrule \midrule
Hearing back: Positive Rating & +2.4\% \\ \hline
Hearing back: Reject          & -5.5\% \\ \hline
Job Impression                & +8.7\% \\ \hline
Sessions                      & +0.29\% \\
\bottomrule
\end{tabular}
\label{tab.taj}
}
\end{table}

\subsection{Discussion}
\label{sec.discussion}

\paragraph{\textbf{GNN in LLM v.s. LLM in GNN}} Recently, some researchers 
\cite{fatemi2023talk,zhu2024understanding} 
propose to encode graph-structured data as prompt and pass into LLM models to complete some reasoning and prediction tasks. The process is generally split into two phases: (1) graph encoding; (2) graph prompt engineering. The encoding step is to convert graph-structured information into a sequence (of language) that can be consumed by language models. Graph prompt engineering is to find the correct way to phrase the question and prediction task that an LLM model is able to return the corresponding answer. The limitations comparing with consuming LLM in GNN training are: (1) The scalability is one of the major bottlenecks. Encoding the graph information may work with small graph, but in industry, if the degree of nodes is above hundreds, encoding graph in LLM may not be scalable, and also may result in huge latency in online serving. (2) Prompt engineering is generally case-specific and model-dependent, making it difficult to scale across multiple business scenarios -- something that \textit{STAR} is designed to handle effectively.

\paragraph{\textbf{LLM v.s. LLM in GNN}} 

LLM embeddings can also be incorporated directly into ranking models. As discussed in Section~\ref{sec.methodology_llm}, optimal performance with LLM embeddings is achieved when they have a large dimensionality -- 4096 in our case -- to capture nuanced semantic meanings. Compared to GNN embeddings, which can be very effective with a dimensionality of 200, LLM embeddings present an additional latency challenge for the online infrastructure, as shown in Table~\ref{tab.latency}, where adding LLM significantly increases latency. Another important consideration is that LLM embeddings do not capture relational information effectively. In future work, we aim to reduce the dimensionality of LLM embeddings to enable efficient integration with \textit{STAR} and direct use in ranking models.

\begin{table}
\small{
\caption{Online latency of LLM and GNN embeddings. }
\vspace{-6pt}
\begin{tabular}{@{\hskip 8pt}l@{\hskip 8pt}|l@{\hskip 8pt}|l}
\toprule
Metric Name &  +LLM embeddings & +GNN embeddings \\
\midrule \midrule
Job Recommendation $P50$    & +5.3\%  & +0.5\%  \\ \hline
Job Search $P50$  & +4.4\%  & +0.8\% \\
\bottomrule
\end{tabular}
\label{tab.latency}
}
\end{table}

\paragraph{\textbf{Co-train v.s. Train Separately}} As we discussed in the Section~\ref{sec.related_work}, some researchers has been working on the co-training and fine-tuning framework of LLM + GNN  recently. We also spent effort in conducting extensive offline evaluations, and we found there are 2 major issues: (a) Both co-training and fine-tuning impose heavy hardware requirements, which makes them extremely costly. We used Flan-T5-small as an example to initiate co-training. The results show that on a V100 GPU with 230GB of memory, the batch size decreases from 4,096 to 512, while the training time increases from 48 hours to approximately 550 hours; (b) the offline training results of co-training are significantly worse than our approach. We observed that incorporating an LLM into the graph engine makes the training process difficult to converge.

\section{Conclusion}
\label{sec.conclusion}
We proposed the \textit{STAR} system, an efficient, scalable, and easy-to-maintain signal platform that leverages large language models (LLMs) and graph neural networks (GNNs) for feature engineering. \textit{STAR} unifies signals across various LinkedIn products, enhancing the performance of recommendation systems across diverse applications. Results from both offline and online A/B tests demonstrate significant improvements in three different products. 
While we use LinkedIn products as an illustrative example, the solution has a relatively low infrastructure cost and is easy to maintain from a software engineering perspective, making it broadly applicable to most relevance systems in the industry.
We have shared the design and modeling techniques behind this platform, and we believe the design principles and insights will benefit many other ML practitioners working on relevance systems. 
In the future, we aim to further enhance \textit{STAR} by exploring advanced learning algorithms, optimizing for larger datasets, and integrating emerging machine learning technologies.

\appendix
\label{sec.appendix}
\section{Appendix}
\subsection{Evaluation on GNN + LLM Encoder}
\begin{table*}[htbp]
    \centering
    \caption{Performance and resource comparison of GNN+LLM training configurations}
    \scalebox{0.8}{
    \begin{tabular}{l|l|l|l|l|l|l|l|l|l}
    \toprule
        Text Encoder Model & Co-train & Max Sequence Length & Batch Size  & Sampled Neighbors & Learning Rate & Epoch & Val AUC & GPU memory & Training Time \\ \midrule \midrule
        Baseline & / & / & 4096 & 100 & 1.00E-03 & 10 & 0.7324 & 8GB & 48h \\ \hline
        LE-5-xsmall & No & 100 & 512 & 100 & 1.00E-03 & 10 & 0.7175 & 40GB & 140h \\  \hline
        Flan-T5-small & No & 64 & 512 & 100 & 1.00E-04 & 9 & 0.7355 & 78GB & 168h \\  \hline
        Flan-T5-small & Yes & 128 & 96 & 10 & 2.00E-05 & 3 & 0.7233 & 65GB & 168h \\ \hline
        Flan-T5-base & No & 1024 & 64 & 10 & 3.00E-05 & 8 & 0.7172 & 78GB & 168h \\ \hline
        Flan-T5-base & Yes & 1024 & 2 & 10 & 1.00E-06 & 0.1 & \textit{failed} & 65GB & 168h \\ \bottomrule
    \end{tabular}}
    \label{tab:GNN+LLM-encoder}
\end{table*}
\label{sec:gnn+llm-encoder}

We integrated several text encoders, such as Flan-T5-small, Flan-T5-base~\citep{chung2024scaling}, and LE-5-xsmall (an in-house model fine-tuned from the Multilingual-E5-small~\citep{wang2024multilingual} model with a 6-layer encoder on LinkedIn data)—to represent node text features in GNN. We conducted various experiments, comparing freezing the text encoder in the GNN versus co-training the GNN and text encoder. These experiments used both short and long text inputs and explored different hyper-parameters, such as learning rate, batch size, and sampled neighbors. Additionally, we applied various speed up techniques, including LoRA, local SGD, adaptive sequence length within a batch, and Local Gradient Aggregation. Table~\ref{tab:GNN+LLM-encoder} shows the best validation AUC results for different models and setups, along with their respective computation time and resource cost. Please note, at LinkedIn, we limit the duration of a training job to a maximum of 7 days (168 hours) to ensure that no single job monopolizes training infrastructure. With Flan-T5-base co-training enabled, this time constraint allows only 0.1 epochs to be completed within the 168-hour limit.

\begin{table}[htbp]
    \centering
    \caption{Node type distribution in the graph}
    \scalebox{0.9}{
    \begin{tabular}{l|l|l|l|l}
    \toprule
        Node Type & member & job & position & skill \\  \hline
        Count & 376M & 240M & 121M & 42K \\ \midrule \hline
        Node Type & title & company & recruiter & hire project \\ \hline
        Count & 25K & 19M & 596K & 6M \\
        \bottomrule
    \end{tabular}}
    \label{tab:nodes-gnn-llm}
\end{table}

\begin{table}[!ht]
    \centering
    \caption{Edge type distribution in the graph}
     \scalebox{0.7}{
    \begin{tabular}{l|l|l}
    \toprule
        Edge Type & job-title & job-company \\ 
        Count & 417M & 370M \\ \hline
        Edge Type & member-title & member-company \\ 
        Count & 568M & 462M \\ \hline
        Edge Type & member-position (current job) & member-job action (APPLY) \\ 
        Count & 553M & 2,722M \\ \hline
        Edge Type & job-skill & job-position \\ 
        Count & 3,977M & 358M \\ \hline
        Edge Type & member-skill & member-position (past job) \\ 
        Count & 1,247M & 15M \\ \hline
        Edge Type & member-job action (SAVE) & member-recruiter positive interaction \\ 
        Count & 724M & 171M \\ \hline
        Edge Type & recruiter-member & member-hire project  \\ 
        Count & 92M & 582M \\ 
        \bottomrule
    \end{tabular}}
    \label{tab:edges-gnn-llm}
\end{table}

Each experiment was conducted on 12 Nvidia A100 (80GB) GPUs with 50 million training examples and 0.5 million validation examples, using a SAGE encoder \cite{liu2024linksage} dimension of 100.

Our findings revealed trade-offs when using GNNs with LLMs, whether freezing or co-training the text encoder. Freezing the Flan-T5-small encoder resulted in a slight improvement in AUC but required 10 times more training time and significant GPU resources. Co-training GNN with text encoder like Flan-T5-base increased resource consumption without performance gains. This likely stemmed from the need to reduce batch size and the number of sampled neighbors to accommodate the text encoder models and their gradient updates within GPU memory, ultimately hurting performance. Given the high computational cost and marginal improvements, neither freezing nor co-training the text encoder proved practical for efficient use in our case.

\balance

\bibliographystyle{ACM-Reference-Format}
\bibliography{main}

\end{document}